\documentclass[11pt]{article}
\usepackage{geometry}
\usepackage{coling2020}
\usepackage{times}
\usepackage{url}
\usepackage{latexsym}
\usepackage{microtype}
\usepackage{amsmath,amsthm,amsfonts,amssymb,amscd}
\usepackage{verbatim}

\colingfinalcopy 

\title{Reed at SemEval-2020 Task 9: Fine-Tuning and Bag-of-Words Approaches to Code-Mixed Sentiment Analysis}

\author{Vinay Gopalan \\
   Reed College \\
  {\tt gopalanvinay@gmail.com} \\\And
  Mark Hopkins \\
   Reed College \\
  {\tt hopkinsm@reed.edu} \\}

\date{}

\begin{document}
\maketitle
\begin{abstract}
  We explore the task of sentiment analysis on Hinglish 
  (code-mixed Hindi-English) tweets as participants of Task 9 of the SemEval-2020 competition, 
  known as the SentiMix task. We had  two main approaches: 
  1) applying transfer learning by fine-tuning pre-trained BERT models and 2) training feedforward neural networks on 
  bag-of-words representations. During the 
  evaluation phase of the competition, we obtained an F-score of 71.3\% with our best model, which 
  placed $4^{th}$ out of 62 entries in the official system rankings.
\end{abstract}

\section{Introduction}
\label{intro}

%
%
\blfootnote{
    %
    %
    %
    %
    %
    %
    \hspace{-0.65cm}  
    This work is licensed under a Creative Commons 
    Attribution 4.0 International License.
    License details:
    \url{http://creativecommons.org/licenses/by/4.0/}.
}

The Internet today has a vast collection of data in various forms, including text and 
images. A big part of this data comes from various social media platforms, which enable 
users to share thoughts about their daily experiences. The millions of 
tweets, updates, location check-ins, likes, and dislikes that users share every day on different 
platforms form a large bank of opinionated data. Extracting sentiment from this data, though immensely useful, is also challenging, giving rise to the NLP task known as \textit{sentiment analysis}. Several models have been proposed to perform this task over the years, such as 
those built on top of the Recursive Neural Tensor Network \cite{socher2013recursive} 
or the more recent BERT model \cite{devlin2018bert}. Although most of these language 
technologies are built for the English language, a lot of Internet data comes from multilingual speakers who combine 
English with other languages when they use social media. Thus, it is also important to study
sentiment analysis for this so-called `code-mixed' social media text. In this paper, 
we explore sentiment analysis on code-mixed Hinglish (Hindi-English) 
tweets, specifically as a participant in Task 9 of SemEval-2020 \cite{patwa2020sentimix}.

We had two main strategies. We began by fine-tuning 
pre-trained BERT models to our 
target task. During this initial phase, we observed that the accuracies yielded
by \texttt{bert-base}, \texttt{bert-multilingual} and \texttt{bert-chinese} on the validation data 
were approximately the same. This made us hypothesize that the pre-trained weights were mostly unhelpful for the task. To test this, we attempted to recreate the BERT fine-tuning 
results only using feedforward neural networks trained on a 
bag-of-words (BoW) representation.

We found that the the results of fine-tuning \texttt{bert-large} (24 layers) 
could be approximated by a 2-layer BoW feedforward neural network, as our best-performing fine-tuned model had an accuracy of 63.9\% 
and our best-performing bag-of-words model had an accuracy of 60.0\% on the validation corpus. In the evaluation phase of the competition, 
our fine-tuned model had an F-score of 69.9\% without bagging \cite{breiman1996bagging} 
and 71.3\% with bagging. Once the official system rankings were 
published on April 6, 2020\footnote{see https://competitions.codalab.org/competitions/20654\#learn\_the\_details-results}, 
our best submission (CodaLab username: \texttt{gopalanvinay}) placed 4th out of 62 entries. 
All code needed to recreate our results can be found here\footnote{see https://github.com/gopalanvinay/thesis-vinay-gopalan}.

\section{Background}

The organizers of SentiMix simultaneously organized a Hinglish 
and an analogous Spanglish (code-mixed Spanish-English) task \cite{patwa2020sentimix}. We participated in the Hinglish
track.

The task was to predict the sentiment of a given code-mixed tweet. Entrants were 
provided with training and validation data. These datasets were comprised of Hinglish 
tweets and their corresponding sentiment labels: positive, negative, 
or neutral.  Besides the sentiment labels, the organizers also provided language labels at the word level. 
Each word is tagged as English, Hindi, or universal (e.g. symbols, mentions, hashtags). 
Systems were evaluated in terms of \emph{precision}, \emph{recall} and \emph{$F_1$-measure}.

All data was provided in tokenized CoNLL format. In this format, the first line (prefaced with the term ``meta") provides a
unique ID and the sentiment (positive, negative or neutral) of the tweet. Each tweet is then segmented 
into word/character tokens and each token is given a language ID, which is either 
\emph{HIN} for Hindi, \emph{ENG} for English and \emph{O} if the token is in neither 
language. Below is an example for a positive-sentiment tweet with id 173:
\begin{table}[h!] 
  \begin{center} 
  \begin{tabular}{l l l} 
meta&173&positive \\
@&O \\
BeingSalmanKhan&Eng \\
It&Eng \\
means&Eng \\
sidhi&Hin \\
sadhi&Hin \\
ladki&Hin \\
best&Eng \\
couple&Eng \\
  \end{tabular}
  \end{center}
  \label{conllStructure} 
\end{table}

\noindent The data contains tokens like usernames (e.g. @BeingSalmanKhan) and URLs, 
which we deemed unhelpful for sentiment analysis. Prior to any training, we converted the CoNLL format into a simpler format, consisting of the reconstituted tweet (omitting usernames and URLs) and the sentiment. For the above example, the converted tweet would be: 

\begin{quotation}
\textit{It means sidhi sadhi ladki best couple}	\hfill (positive)	
\end{quotation}

\noindent Note that we ignore the token-level language ids.

\section{Experimental Setup}

For the experimental results reported in this paper (except for the final system results, which use the official competition test set), we trained our systems using the provided training set of 14K tweets and report results on the provided validation set of 3K tweets. Since the task is a relatively balanced three-way classification task, we used simple accuracy as our hillclimbing metric.

We used PyTorch v1.2.0
\footnote{see https://pytorch.org/docs/1.2.0/} and Python 3.7.1\footnote{see https://docs.python.org/3/}. 
All additional details needed to replicate our results are provided in the README of our project's code repository
\footnote{see https://github.com/gopalanvinay/thesis-vinay-gopalan}.

\section{Fine-tuned BERT}

For our baseline experiments, we used the PyTorch-based implementation of BERT \cite{devlin2018bert} provided by Huggingface's \texttt{transformers} package \cite{wolf2019transformers}.  We adapted the \texttt{SST-2} task of the \texttt{run\_glue.py} script. The original task was configured for the binary classification of sentences from the Stanford Sentiment Treebank \cite{socher2013recursive}, as administered by the General Language Understanding Evaluation (GLUE) benchmark \cite{wang2018glue}. We adapted the task for three-way (positive, negative, neutral) classification.

\subsection{Experiment: Varying the Base Model}

In our first experiment, we compared several pre-trained models:

\begin{itemize}
	\item \texttt{bert-base-cased}: A 12-layer transformer with token embeddings of size 768, trained by Google on English data.
	\item \texttt{bert-large-cased}: A 24-layer transformer with token embeddings of size 1024, trained by Google on English data.
	\item \texttt{bert-base-multilingual-cased}: A 12-layer transformer with token embeddings of size 768, trained by Google on the Wikipedia dumps from 104 languages, including Hindi and English.
\end{itemize}

\begin{table}[tb] 
  \begin{center} 
  \begin{tabular}{c|c} 
  \textbf{pretrained model} & \textbf{accuracy (\%)}   \\ 
  \hline
  \texttt{bert-base-cased}   &   62.2 \\  
  \texttt{bert-large-cased}    &  63.3            \\ 
  \texttt{bert-base-multilingual-cased} & 62.3    \\
  \texttt{bert-base-chinese} & 61.0    \\
  \end{tabular}
  \end{center}
  \caption{Results from fine-tuning BERT using various pre-trained models.}
  \label{tab:varybase} 
\end{table}

\noindent For these experiments, we used the default parameters provided by the Huggingface training script. Our hypothesis was that the \texttt{bert-base-multilingual-cased} model, which is simultaneously pre-trained on both Hindi and English, would be more effective\footnote{In retrospect, this was perhaps a naive hypothesis, given that the Hindi-English tweets are in a romanized alphabet, whereas the Hindi used for pre-training \texttt{bert-base-multilingual-cased} was presumably written mostly in Devanagari script.} for Hindi-English tweet classification, but this did not turn out to be the case. In fact, there was little difference between the English-only \texttt{bert-base-cased} model and the \texttt{bert-base-multilingual-cased} model.

This observation suggested a new hypothesis, which was that pretraining was possibly not at all helpful for the code-mixed domain, and that the Transformer was simply learning to classify ``from scratch," starting from the unhelpful weight initializations provided by the pretrained models. To test this hypothesis, we performed fine-tuning with the \texttt{bert-base-chinese} model, a 12-layer transformer with token embeddings of size 768, trained by Google on traditional and simplified Chinese text. The final line of Table~\ref{tab:varybase} shows this result, which is surprisingly close to the models trained using English data, suggesting that the pre-training has relatively little impact on the task performance.

\section{Bag-of-Words Models}

\newcommand*{\field}[1]{\mathbb{#1}}%

The fine-tuning results suggested two possibilities:

\begin{enumerate}
	\item Even though the pre-training makes little difference, the complex Transformer model still learns deep and interesting patterns to achieve an accuracy in the low sixties.
	\item The Transformer model is only learning simple heuristics that could be just as easily learned by simpler models.
\end{enumerate}

\noindent To distinguish between these possibilities, we attempted to replicate the performance of the fine-tuned systems using classical bag-of-words (BoW) models \cite{mctear2016conversational}.

To create a BoW system, we iterated through the training data and stored
the frequency of each word in the corpus. We then created a 
\textit{vocabulary} $\textbf{V} = \{v_{1}, v_{2}, ..., v_{n}\}$, which 
was the set of words that appeared with frequency at least $K$ in the training data, 
for some positive \emph{frequency threshold} $K \in \field{N}$. 
After removing stop words from the vocabulary, we used this 
vocabulary to transform each tweet into an $n$-length vector whose $j^{th}$ element equaled 1 if word $v_{j}$ $\in \textbf{V}$ appeared in the tweet (and 0 otherwise).

We then trained a simple classifier using these BoW vector representations of the tweets. For our classifier, we used a 
standard feedforward neural network with $M$ layers and hidden size $H$. We built and 
trained the NNs using PyTorch \cite{NEURIPS2019_9015}. In order to classify tweets into 
positive, negative and neutral, the final layer applies the softmax function to the 
3-dimensional output of the NN, which normalizes the output into a probability 
distribution over the three possible sentiments of the input tweet. We used a cross-entropy 
loss function for training.

\subsection{Variant: Count-of-Words}
We also experimented with 
a count-of-words representation, where instead of just representing whether a vocabulary 
word appears in a particular tweet as a binary value, we instead represent the 
\textit{frequency} of that word in the tweet. The motivation was that the classifier might gain insight into the 
sentiment of the overall sentence based on the frequency of  
`positive' or `negative' vocabulary members.

\subsection{Variant: Bag-of-Ngrams}
We also experimented with a ``bag-of-ngrams" approach that extended our 
vocabulary by including ngrams up to length $N$ (again using a minimum frequency threshold of $K$). 
Through this extension, we hoped the trained systems could exploit  contextual information from neighboring words that are more than the sum of their parts, like ``pretty good" or ``awfully well done."

\subsection{Experiments}

	\begin{table}[tb!] 
			\caption{Experimental results for our bag-of-words models, varying the frequency threshold $K$. These experiments each use a two-layer network with a hidden size $H=784$.\label{bowresults1}}  
			\begin{center} 
			\begin{tabular}{c c} 
			\hline
			Frequency Threshold (K) &  Accuracy \\ 
			\hline
				10 & 58.8\%	 \\ 
				15 & 58.6\%	 \\ 
				20 & 58.3\%	 \\ 			
			\hline
			\end{tabular}
			\end{center}		
		\end{table}
				
		\begin{table}[tb!] 
			\caption{Experimental results for our bag-of-words models, varying the number of layers of the feedforward neural classifier. These experiments each use the same frequency threshold $K=15$. 
			\label{bowresults2}}  
			\begin{center} 
			\begin{tabular}{c c c} 
			\hline
			Number of NN Layers & Hidden Layer Size (H) & Accuracy \\ 
			\hline
				2	& 300 & 59.3\%	 \\ 			
				2	& 768 & 58.6\%	 \\ 
				3	& 768 & 58.0\%	 \\ 
				4	& 768 & 57.8\%	 \\
			\hline
			\end{tabular}
			\end{center}
		\end{table}

		\begin{table}[tb!] 
			\caption{Experimental results for our bag-of-ngrams models using unigrams, bigrams, and trigrams, using a frequency threshold $K=15$ for all ngrams. These experiments use 2-layer feedforward neural networks with a hidden layer size $H=300$. \label{bowresults3}} 
			\begin{center} 
			\begin{tabular}{c c} 
			\hline
			\textbf{Ngrams Used} & \textbf{Accuracy} \\ 
			\hline
				uni		& 59.3\%	 \\ 
				uni + bi	& 60.0\%	 \\ 
				uni + bi + tri & 60.0\% \\
			\hline
			\end{tabular}
			\end{center}
			
		\end{table}

For our experiments, we varied 3 hyperparameters: the frequency threshold $K$ for the words/ngrams (Table~\ref{bowresults1}), as well as the number of NN layers and hidden layer size $H$(Table~\ref{bowresults2}). 

Unsurprisingly, system accuracy improved (Table~\ref{bowresults1}) as we decreased the frequency threshold $K$. This at least confirmed the natural hypothesis 
that a larger vocabulary size would lead to more information and improved sentiment analysis. 

Increasing the number of layers was somewhat detrimental to model performance, as was widening the hidden layer size (Table~\ref{bowresults2}). Perhaps this was caused by overfitting and might have been fixed by a regularizer, but nevertheless we stuck to simple two-layer networks for the remainder of the experiments.

After this first round of experiments, we conducted additional experiments using the count-of-words 
approach.  
In general, the accuracies in the count-of-words approach were 
slightly lower than the simple BoW approach.

For our bag-of-ngrams experiments (Table~\ref{bowresults3}), the inclusion of bigrams improved the system results, while the addition of trigrams did not have much impact, likely because of the absence of sufficiently frequent trigrams (around 10, using a frequency threshold $K=15$). 

\subsection{Takeaways}

With only minor hillclimbing effort, we were able to train a simple bag-of-bigrams classifier to a validation accuracy of $60\%$. Contrast this to the validation accuracy of $62.2\%$ achieved via fine-tuning of \texttt{bert-base-cased}, and the validation accuracy of $61.0\%$ achieved via fine-tuning of \texttt{bert-base-chinese}. The similarity of these results suggests that the success of the BERT models is probably not due to a deep understanding of the code-mixed language, but rather the Transformer's ability to exploit simple word count statistics slightly better than a simple bag-of-words classifier.

\section{Final System Submissions}

Because we obtained slightly better performance with the fine-tuned BERT model, we used this as the basis for our competition system. To improve its performance, we experimented\footnote{We also experimented with the \emph{roberta-base} model \cite{liu2019roberta}, but were not able to improve of BERT's results.} with various hyperparameter settings, including \emph{learning rate}, 
\emph{weight decay}, \emph{max\_grad\_norm} and \emph{Adam epsilon}. Our best validation result of 63.8\% was yielded by \emph{bert-large-cased} using a 
learning rate of $2\times10^{-5}$,  a weight decay of 0, a \emph{max\_grad\_norm} of 1, and an Adam epsilon parameter of $1\times10^{-8}$. This system became our first submitted system, achieving an F-score of 69.9\% according to the official scoring script.

For our second submission, we used bagging \cite{breiman1996bagging} to make our BERT classifier more robust. We created 10 bootstrap samples from the given training data. We then trained 10 instances of the \texttt{bert-large-cased} model on these 
bootstrap samples, and then combined the results of all the models 
by voting, i.e., if a plurality of the 10 models predicted a particular tweet as 
positive (negative/neutral), then the tweet’s label is deemed as positive (negative/neutral). Our bagged system obtained an 
F-score of 71.3\%, placing us $4^{th}$ overall in the competition out of 62 entrants.

\section{Conclusion}

  In this paper we investigated sentiment analysis on code-mixed Hinglish 
  (Hindi-English) tweets as participants of Task 9 of the SemEval 2020 competition. 
  We implemented two main approaches: 
  1) applying transfer learning by fine-tuning pre-trained models like BERT and 2) training feedforward neural networks on 
  bag-of-words representations. We found that the results of fine-tuning \texttt{bert-large-cased}
  (24 layers) could be approximated by a 2-layer BoW feedforward NN. 
  During the evaluation phase of the competition, our top system obtained an F-score of 71.3\%, placing $4^{th}$ out of 62 entries in the official system rankings.
  
  The fact that we managed fourth place with a system that was little better than a bag-of-words classifier suggests that existing pre-trained vector representations are not particularly good models of code-mixed language. One obvious reason is the difference between the pre-training domain (English Wikipedia) and the target domain (code-mixed tweets). While we were initially hopeful that the multilingual BERT model released by Google might be more effective than the English-only models, it was not. It remains unresolved whether this is because of the formality differences between the two domains, because of the lack of romanized Hindi in Google's training data, or because language models simultaneously pre-trained on multiple languages do not generalize sufficiently to properly understand code-mixed data.

\bibliographystyle{coling}
\bibliography{semeval2020}

\begin{thebibliography}{}

\bibitem[\protect\citename{Breiman}1996]{breiman1996bagging}
Leo Breiman.
\newblock 1996.
\newblock Bagging predictors.
\newblock {\em Machine learning}, 24(2):123--140.

\bibitem[\protect\citename{Devlin \bgroup et al.\egroup }2018]{devlin2018bert}
Jacob Devlin, Ming-Wei Chang, Kenton Lee, and Kristina Toutanova.
\newblock 2018.
\newblock Bert: Pre-training of deep bidirectional transformers for language
  understanding.
\newblock {\em arXiv preprint arXiv:1810.04805}.

\bibitem[\protect\citename{Liu \bgroup et al.\egroup }2019]{liu2019roberta}
Yinhan Liu, Myle Ott, Naman Goyal, Jingfei Du, Mandar Joshi, Danqi Chen, Omer
  Levy, Mike Lewis, Luke Zettlemoyer, and Veselin Stoyanov.
\newblock 2019.
\newblock Roberta: A robustly optimized bert pretraining approach.
\newblock {\em arXiv preprint arXiv:1907.11692}.

\bibitem[\protect\citename{McTear \bgroup et al.\egroup
  }2016]{mctear2016conversational}
Michael~Frederick McTear, Zoraida Callejas, and David Griol.
\newblock 2016.
\newblock {\em The conversational interface}, volume~6.
\newblock Springer.

\bibitem[\protect\citename{Paszke \bgroup et al.\egroup
  }2019]{NEURIPS2019_9015}
Adam Paszke, Sam Gross, Francisco Massa, Adam Lerer, James Bradbury, Gregory
  Chanan, Trevor Killeen, Zeming Lin, Natalia Gimelshein, Luca Antiga, Alban
  Desmaison, Andreas Kopf, Edward Yang, Zachary DeVito, Martin Raison, Alykhan
  Tejani, Sasank Chilamkurthy, Benoit Steiner, Lu~Fang, Junjie Bai, and Soumith
  Chintala.
\newblock 2019.
\newblock Pytorch: An imperative style, high-performance deep learning library.
\newblock In {\em Advances in Neural Information Processing Systems 32}. Curran
  Associates, Inc.

\bibitem[\protect\citename{Patwa \bgroup et al.\egroup
  }2020]{patwa2020sentimix}
Parth Patwa, Gustavo Aguilar, Sudipta Kar, Suraj Pandey, Srinivas PYKL,
  Bj{\"o}rn Gamb{\"a}ck, Tanmoy Chakraborty, Thamar Solorio, and Amitava Das.
\newblock 2020.
\newblock Semeval-2020 task 9: Overview of sentiment analysis of code-mixed
  tweets.
\newblock In {\em Proceedings of the 14th International Workshop on Semantic
  Evaluation ({S}em{E}val-2020)}, Barcelona, Spain, December. Association for
  Computational Linguistics.

\bibitem[\protect\citename{Socher \bgroup et al.\egroup
  }2013]{socher2013recursive}
Richard Socher, Alex Perelygin, Jean Wu, Jason Chuang, Christopher~D Manning,
  Andrew~Y Ng, and Christopher Potts.
\newblock 2013.
\newblock Recursive deep models for semantic compositionality over a sentiment
  treebank.
\newblock In {\em Proceedings of the 2013 conference on empirical methods in
  natural language processing}, pages 1631--1642.

\bibitem[\protect\citename{Wang \bgroup et al.\egroup }2018]{wang2018glue}
Alex Wang, Amanpreet Singh, Julian Michael, Felix Hill, Omer Levy, and Samuel~R
  Bowman.
\newblock 2018.
\newblock Glue: A multi-task benchmark and analysis platform for natural
  language understanding.
\newblock {\em arXiv preprint arXiv:1804.07461}.

\bibitem[\protect\citename{Wolf \bgroup et al.\egroup
  }2019]{wolf2019transformers}
Thomas Wolf, Lysandre Debut, Victor Sanh, Julien Chaumond, Clement Delangue,
  Anthony Moi, Pierric Cistac, Tim Rault, R{\'e}mi Louf, Morgan Funtowicz,
  et~al.
\newblock 2019.
\newblock Transformers: State-of-the-art natural language processing.
\newblock {\em arXiv preprint arXiv:1910.03771}.

\end{thebibliography}

\end{document}